\documentclass[11pt,a4paper]{article}
\usepackage{amsmath,graphicx}
\usepackage[hyperref]{acl2021}
\usepackage{times}
\usepackage{latexsym}

\usepackage{microtype}

\aclfinalcopy 

\title{SGD-QA: Fast Schema-Guided Dialogue State Tracking for Unseen Services}

\author{Yang Zhang~~~~ Vahid Noroozi~~~~ Evelina Bakhturina~~~~ Boris Ginsburg \\
  
  NVIDIA, CA, USA \\
  \texttt{\{yangzhang, vnoroozi, ebakhturina, bginsburg\}@nvidia.com}
  }

\date{}

\begin{document}
\maketitle
\begin{abstract}
Dialogue state tracking is an essential part of goal-oriented dialogue systems, while most of these state tracking models often fail to handle unseen services.
In this paper, we propose SGD-QA, a simple and extensible model for schema-guided dialogue state tracking based on a question answering approach.
The proposed multi-pass model shares a single encoder between the domain information and dialogue utterance. The domain's description represents the query and the dialogue utterance serves as the context. The model improves performance on unseen services by at least 1.6x compared to single-pass baseline models on the SGD dataset. SGD-QA shows competitive performance compared to state-of-the-art multi-pass models while being significantly more efficient in terms of memory consumption and training performance.
We provide a thorough discussion on the model with ablation study and error analysis.
\end{abstract}

\section{Introduction}

\label{sec:intro}


Goal-oriented dialogue systems are designed to help a user solve a specific task, e.g. making a hotel reservation\cite{zhang2020recent}. Dialogue state tracking (DST), at the core of dialogue systems, extracts the relevant semantic information from user-system dialogues such as mentioned entities, called \emph{slots}. Usually, dialogue state tracking models consist of (1) a neural natural language understanding (NLU) module, which identifies slots and their values from a dialogue turn, and (2) a simple, often rule-based, state tracker (ST), which builds up the dialogue state over the course of the dialogue.

We can categorize NLU models proposed for schema-guided datasets into single-pass~\cite{rastogi2019towards,balaraman2020domain, noroozi2020fastsgt} and multi-pass~\cite{li2020sppd,ruan2020fine,xingshi2020,ma2019end,junyuan2020} models.
Single-pass models encode dialogue information once per turn to predict all occurring entities. This entails that the model's architecture needs to be ontology-specific, because it needs to predict the value of every possible slot at the output decoding layers. This approach neither scales with growing ontology nor generalizes well to new domains.

Multi-pass models run the encoder for each possible slot and value, predicting only a score for each pass for how likely this input entity occurs in the dialogue turn. These models turn out to generalize much better on new domains since they transfer and share knowledge between different services, domains and slots. One of the state-of-the-art multi-pass models is SPPD model~\cite{li2020sppd}. It has shown great performance is terms of accuracy, although it uses multiple models, and thus is less memory and computational efficient its single-pass counterparts.

We propose a multi-pass model SGD-QA inside the NVIDIA NeMo toolkit\footnote{ https://github.com/NVIDIA/NeMo}~\cite{kuchaiev2019nemo}, which is a simple and extensible BERT-based~\cite{devlin2019bert} NLU model for dialogue state tracking on the The Schema-Guided Dialogue (SGD) dataset~\cite{rastogi2019towards} that outperforms significantly single-pass models like SGD baseline model~\cite{rastogi2019towards} or FastSGT~\cite{noroozi2020fastsgt} on unseen services. SGD-QA is ontology-agnostic, which makes it applicable to new services and domains. Unlike the multi-pass model SPPD~\cite{li2020sppd}, we show it is sufficient to use a single BERT model, significantly reducing memory consumption. Despite being a multi-pass model, SGD-QA is optimized to converge at comparable time as its single-pass counterparts. At the same time, it can be parallelized during inference for low latency.

\section{Related Works}
\label{sec:related_work}

Most of the existing public dialogue datasets, such as MultiWOZ~\cite{budzianowski2018multiwoz} and Machines Talking To Machines~\cite{shah2018building}, use a fixed list of predefined slots for each domain without any information on their semantics. As a result, information can not be shared between similar domains nor slots.

The Schema-Guided Dialogue (SGD) dataset\footnote{ https://github.com/google-research-datasets/dstc8-schema-guided-dialogue}~\cite{rastogi2019towards} was created to overcome these challenges and facilitate models with zero-shot capabilities. It is the biggest dataset for goal-oriented dialogue state tracking with 20k annotated dialogues for 45 services spanning 20 domains.
The SGD dataset defines an ontology, called \emph{schema}, that contains descriptions in natural language for all entities associated with a particular service.  Slots are further classified into non-categorical slots and categorical slots. For categorical slots, the schema also includes a list of possible values. The user-system dialogues can be either single-domain or multi-domain, where a user can request two or more services per dialogue. Each turn is labeled with relevant schema information, called dialogue \emph{state}, comprising of the active intent, requested user slots and slot assignments that occurred throughout the dialogue.
The SGD dataset is designed to test services beyond those seen at training: 57\% of the dev and 78\% of the test dataset stem from with unseen services. Nevertheless, seen and unseen services can share similar slots and functionality.


\begin{figure*}[ht!]
 \centering
 \includegraphics[width=0.7\textwidth,page=4,trim=0 6.8in 0 0.72in,clip]{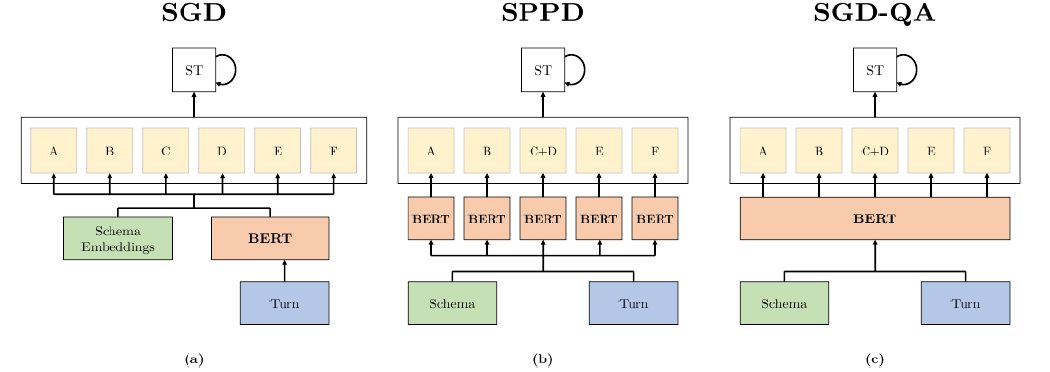}
 \caption{\label{fig:models} Model comparison between SGD baseline model, SPPD and SGD-QA. "ST" is an abbreviation for state tracker, which can include additional improvements that differ between the models. A, B, C, D, E, F stand for the tasks: user's intent prediction, requested slots prediction, categorical status prediction, non-categorical status prediction, categorical slot value prediction and non-categorical slot value extraction. The input is a combination of schema information and dialogue turn. For the SGD baseline model, the schema embeddings are non-trainable.}
\end{figure*}

\section{SGD-QA Model}
\label{sec:model}
 SGD-QA is a multi-pass NLU model that uses the same input representation as SPPD~\cite{li2020sppd}, while training a single BERT~\cite{devlin2019bert} encoder with multi-task heads like the SGD baseline model. 
We use a pre-trained BERT model with four classification heads and one span prediction head for slot value extraction.
The sequence classification heads are comprised of three linear layers with Tanh and GELU~\cite{hendrycks2016gaussian} activation functions in between. 
Similar to SPPD, the SGD-QA model relies on a question answering approach and uses schema description as query and the dialogue turn as context. Each input instance is intended for exactly one task which is implemented by masking out the other task's losses. The final loss is the average of all losses.
Table \ref{tab:input} shows the input format into BERT encoder for each of the tasks.

\begin{table}[!b]
\centering
\resizebox{\columnwidth}{!}{
\begin{tabular}{c|c|c}
 \textbf{Task} & \textbf{Sequence 1} & \textbf{
 Sequence 2}\\ \hline
 Intent  & intent name \&  description & sys+usr utt \\
 Slot request  & slot name \&  description & usr utt \\
 Slot status & slot name \&  description & sys+usr utt  \\
 Value predication &  slot name \&  value name & sys+usr utt \\
 Value extraction &  slot name \&  description & sys+usr utt 
\end{tabular}}
\caption{\label{tab:input} Input sequences for the pre-trained BERT model for each task. Sequence 2 is composed of either system or user utterance. Sequence 1 is preceded by "[CLS]" token and separated from sequence 2 with a "[SEP]" token.}
\end{table}

\subsection{Tasks}
Intent prediction, requested slot prediction and categorical slot value prediction are formulated as binary sequence classification.
The slot status prediction task is a 3-way sequence classification to predict whether a slot is active, non-active ("none") or does not matter to the user ("dontcare"). Multiple slots can become active in a single turn. 
Non-categorical slot value extraction extracts the values for non-categorical slots by using a span-based prediction head, similar to SQuAD~\cite{rajpurkar2016squad}, where two token classification layers detect the start and end positions of the slot value in the dialogue turn. Like the categorical slot value prediction task, its prediction is only used when the slot status is active.

\subsection{State Tracker}
\label{ssec:state_tracker}

The state tracker takes predictions from the NLU model and previous dialogue state to build the next dialogue state, see Figure~\ref{fig:models}.
The basic state tracker as used in the SGD baseline model~\cite{rastogi2019towards} decides based on the slot status whether to carry over a slot value from the previous state, ignore it or derive the value from the slot value prediction or extraction task. FastSGT~\cite{noroozi2020fastsgt}, SPPD~\cite{li2020sppd}, and SGP-DST~\cite{ruan2020fine} all use different state tracking enhancements on top of that.  While state tracking enhancement are not focus of this paper, we do acknowledge they lead to big improvements in goal accuracy, as shown in Table \ref{tab:single_domain}, \ref{tab:all_domain}.


\section{Experiments}
\label{sec:experiments}
We conducted all experiments on NVIDIA DGX1 systems with 8x V100 GPUs. We used pre-trained BERT-base-cased~\cite{devlin2019bert} provided by HuggingFace~\cite{wolf2019huggingfaces} and  WordPiece Tokenizer~\cite{wu2016googles} for input tokenization. We used multi-GPU training with automatic mixed precision~\cite{micikevicius2017mixed} and an effective batch size of 1024 - larger than previous models. Hyperparameters and model parameters can be found in Section \ref{sec:hyperparameters}.
We use $\text{MODEL}$ to denote a model with basic state tracker and $\text{MODEL}_{ST}$ for a model with state tracker enhancements. Code will be released after publication. 


\subsection{Data Balancing}
\label{ssec:data_balancing}


Typically, classification problems face data imbalance within and between tasks. All four classification tasks have more than 72\% negative examples. The value prediction and extraction tasks only make up 4\% and 2\% of the total training data respectively, whereas slot status prediction constitutes 31\%.
To level out this discrepancy between tasks, we balance the slot status prediction data by upper bounding the negative examples by the number of positive examples for that slot, while retaining at least one. 

\subsection{Evaluation Metrics}
Models trained and evaluated on the SGD dataset~\cite{rastogi2019towards} report the evaluation metrics: intent accuracy, slot request F1 score, average goal accuracy and joint goal accuracy. The average goal accuracy denotes overall average accuracy of slot assignments for a dialogue turn. Joint goal accuracy (Joint GA) measures the accuracy of predicting \textit{all} slot assignments for a turn correctly, thus it is the strictest and most cited of them all.

\subsection{Performance Evaluation}
\label{ssec:evaluation}

In this section, we evaluate and compare SGD-QA with the single-pass models SGD baseline model~\cite{rastogi2019towards} and FastSGT~\cite{noroozi2020fastsgt} and also with multi-pass models SPPD~\cite{li2020sppd} and SGP-DST\cite{ruan2020fine}.
The models are evaluated on single-domain and all-domain dialogues of the SGD dataset~\cite{rastogi2019towards}.

On single domain SGD-QA achieves a Joint GA of 65\%, 1.3x higher than FastSGT and SGD baseline model on dev data. 
This performance improvement is largely due to the SGD-QA's generalization power on unseen services, where the Joint GA is 1.8x higher. We think it is because we use a question answering approach and a single BERT for all tasks, which allows information to be shared between schema and dialogue. 
With state tracking enhancements $\text{SGD-QA}_{st}$ has an overall Joint GA 1.4x higher than $\text{FastSGT}_{st}$, with an even higher improvement on unseen services.

On all-domain SGD-QA achieves a Joint GA of 52.3\%, 1.2x higher than SPPD and 1.3x higher than and SGP-DST, see Table \ref{tab:all_domain}. SGD-QA performs 1.6x higher on unseen services than the best single-pass model.
This shows that a single encoder is sufficient for a good NLU model, and we do not need to train separate models for each task like SPPD and SGP-DST, thus reducing memory consumption to $\frac{1}{5}$ and $\frac{1}{6}$ respectively. 
$\text{SGD-QA}_{st}$ with state tracker enhancements performs worse than its multi-pass counterparts.
However, since SPPD and SGP-DST use different enhancements and neither released code, we could not reproduce their state tracker for comparison.
Despite big improvements for multi-pass models by using state tracker enhancements, FastSGT was only able to increase Joint GA from 42\% to 51.6\% due to low performance on unseen services, which shows the limitation of the single-pass model and the importance of the question answering approach for the NLU model which is orthogonal and complementary to the state tracker. Table \ref{tab:dataset} shows the dataset statistics.


SGD-QA uses 3 epochs for training compared to 80 epochs for SGD baseline model and 160 epochs for FastSGT. Due to the shared weights between schema and dialogue encoding the training signal is much stronger, so a fraction of the single-pass models' epochs suffices for convergence. Therefore, the overall training time is comparable to that of a single-pass model. 
At inference and in production, the latency can be reduced by parallelizing the queries since they are conditionally independent.

By switching out BERT-base-cased with BERT-large-cased the Joint GA improved from 74.6\% to 78.3\% on single-domain, see Table \ref{tab:ablation}.
Evaluation on single-domain dialogues can be improved by training all-domain data. This can be considered as data augmentation. This increases the overall Joint GA from 74.6\% to 82.9\%, with most of the improvements stemming from unseen services, see Table \ref{tab:ablation}. The model architecture and memory consumption did not change by switching from single-domain to multi-domain data, which allows SGD-QA to be used for transfer learning.


\begin{table}[t]
\centering
\resizebox{\columnwidth}{!}{
\begin{tabular}{l|cccc}
 \textbf{Model}    & \textbf{Intent Acc}&\textbf{Slot Req F1}&\textbf{Average GA} & \textbf{Joint GA} \\ \hline
 SGD baseline\cite{rastogi2019towards}  & 96.3(99.0/94.2) & 96.6(99.6/94.3) & 76.3(90.8/65.0) & 49.2(71.1/32.4) \\
 FastSGT\cite{noroozi2020fastsgt}  & 96.2(98.8/93.4) & 96.8(99.6/94.6) & 76.6(90.9/69.0) & 49.0(71.0/32.0) \\
 SGD-QA & 97.5(98.8/96.5) & 98.3(99.4/97.5) &  89.5(91.4/88.1) & 65.0(73.1/58.8) \\ \hline
 $\text{FastSGT}_{ST}$  & 96.8(98.8/95.2) & 96.4(99.6/94.0) & 80.1(96.5/68.1) & 56.1(88.0/31.5)\\
 $\text{SGD-QA}_{ST}$ & \textbf{97.9(99.0/96.9)} & \textbf{98.6(99.6/97.9)} & \textbf{95.1(97.6/93.1)} & \textbf{78.3(91.3/68.2)} \\ \hline
\end{tabular}}
\caption{\label{tab:single_domain} Accuracy comparisons between the SGD baseline, FastSGT, and SGD-QA on \textbf{single-domain} dev set for all services (seen/unseen). We excluded the configurations that are not available. Average of three runs are reported.}
\end{table}

\begin{table}[t]
\centering
\resizebox{\columnwidth}{!}{
\begin{tabular}{l|cccc}
 \textbf{Model}   & \textbf{Intent Acc}&\textbf{Slot Req F1}& \textbf{Joint GA} \\ \hline
 
 SGD baseline\cite{rastogi2019towards}
 & 91.0(96.1/87.3) & 97.5(99.6/96.0) & 42.7(\textbf{61.2}/29.3) \\ 
 
 FastSGT\cite{noroozi2020fastsgt} 
 & 90.6(96.2/86.4) & 97.3(99.6/95.5) & 42.0(60.8/28.1) \\
 
 SGP-DST\cite{ruan2020fine}
 & 95.3(95.7/94.8) & 98.4(98.5/98.3) & 39.4(41.7/36.4) \\
 
 SPPD\cite{li2020sppd}
 & 98.5(98.4/98.6) & 99.0(99.7/98.0) & 41.9(41.9/41.9) \\
 
 SGD-QA
 & \textbf{97.3(98.0/96.5)} & \textbf{99.2(99.7/98.5)} & \textbf{52.3}(57.5/\textbf{45.7}) \\ \hline
 
 $\text{FastSGT}_{ST}$
 & 90.5(96.3/86.3) & 97.3(99.7/95.6) & 51.6(79.6/31.4) \\ 
 
 $\text{SGP-DST}_{ST}$
 &95.3(95.7/94.8) & 98.4(98.5/98.3)  & 80.0(88.3/69.2) \\
 
 $\text{SPPD}_{ST}$
 & \textbf{98.5(98.4/98.6)} & 99.0(99.7/98.0) & \textbf{84.0(88.0/78.9)} \\
 

 $\text{SGD-QA}_{ST}$
 & 97.5(98.1/96.7) & \textbf{99.2(99.7/98.5)} & 60.0(66.4/50.8) \\ \hline
 
\end{tabular}}

\caption{\label{tab:all_domain} Accuracy comparisons between SGD baseline, FastSGT, SPPD, SGP-DST, SGD-QA on \textbf{all-domain} dev set for all services (seen/unseen). Average of three runs are reported. }
\end{table}

\begin{table}[ht!]

\centering
\resizebox{\columnwidth}{!}{
\begin{tabular}{l|cccc}
 \textbf{Model}   & \textbf{Intent Acc}&\textbf{Slot Req F1} &\textbf{Joint GA} \\ \hline
 \textbf{SINGLE DOMAIN} \\
 
 SGD-QA unbalanced
 & 97.5(98.8/96.5) & 98.3(99.4/97.5)  & 65.0(73.1/58.8) \\
 
 $\text{SGD-QA}_{ST}$ unbalanced
 & 98.0(99.1/97.1) & 98.7(99.5/98.0) & 74.7(88.2/64.2) \\
 
 $\text{SGD-QA}_{ST}$
 & 98.3(99.1/97.2) & 98.3(99.3/97.5) & 74.6(88.0/64.2) \\
 
 $\text{SGD-QA}_{ST}$ trained on all-domains 
 & \textbf{98.4(99.3/97.4)} & \textbf{99.0(99.7/98.4)}  & \textbf{82.9(92.5/71.5)} \\
 
 $\text{SGD-QA}_{ST}$ Large BERT
 & 97.9(99.0/96.9) & 98.6(99.6/97.9) & 78.3(91.3/68.2) \\
\textbf{ALL DOMAIN} \\
SGD-QA
 & 97.3(98.0/96.5) & 99.2(99.7/98.5)  & 52.3(57.5/45.7) \\
 
 $\text{SGD-QA}_{ST}$ unbalanced
 & 97.6(97.9/97.2) & 99.3(99.7/98.7) & 34.0(36.2/31.1) \\

 $\text{SGD-QA}_{ST}$ 
 & \textbf{97.5(98.1/96.7)} & \textbf{99.2(99.7/98.5)} & \textbf{60.0(66.4/50.8)} \\ \hline
\end{tabular}}

\caption{\label{tab:ablation} Ablation on single-domain dialogues w.r.t. different accuracy metrics on dev set for all services (seen/unseen). Average of three runs.}
\end{table}

\subsection{What did not Work}

Similar to SGP-DST~\cite{ruan2020fine}, we changed the categorical slot value prediction to a value extraction task and fused it with the non-categorical slot value prediction. We hoped both tasks can share information. However, it deteriorated the categorical slot value accuracy. We suspect this is due to the fact that an extractive task is generally harder than a classification task.
    
Using separate BERTs like SPPD~\cite{li2020sppd} model did not improve accuracy. This shows that a single BERT is capable enough to learn all tasks jointly.

Probabilistic averaging is a method used for SPPD. It is a decision tree heuristic to solve ambiguities in the two-stage slot assignment process by considering both slot status and value prediction alike instead of the letting the first being independent from the second.  Unfortunately, we could not find a threshold that would improve the overall goal accuracy.
  
We also tried different data pre-processing changes, including normalizing schema input. Intent names often use snake case, e.g. "FindRestaurant", and slot names use camel case, e.g. "number\_of\_seats". We normalized the names by breaking them into separate words "find restaurant" and "number of seats". We also converted numerical values for categorical slots into string format.

\section{Conclusions}
\label{sec:conslusion}

We proposed SGD-QA, a schema-guided dialogue state tracking model based on a question answering approach. It uses a \emph{single} BERT encoder shared between domain information and dialogue input, which significantly reduces the computation and memory consumption, important in many real-world applications. Our model is consistently better than competitive single-pass models, particularly on unseen services due to its zero-shot capability.
SGD-QA has a simple, ontology-agnostic and modular architecture. We provided error analysis and ablation study. Despite being a multi-pass model, the training and inference processes are very fast, since the model requires fewer training epochs and the inference can be parallelized.

\section*{Acknowledgments}

We thank Miao Li, Abhinav Rastogi and Xiaoxue Zang for their help on the SPPD and SGD baseline model, respectively.

\bibliographystyle{acl_natbib}
\bibliography{acl2021}
\clearpage

\appendix
\section{Hyperparameters}\label{sec:hyperparameters}
SGD-QA model based on BERT-base-cased has 111M parameters, 108M of which are in BERT encoder.
The based on BERT-large-cased has 339M parameters, 333M of which are in BERT encoder.
Table \ref{tab:hyper_parameters} shows the hyperparameters. We tuned learning rate, number of epochs using uniform sampling, and the remaining hyperparameters using manual tuning. 

\begin{table}[h]

\centering
\resizebox{\columnwidth}{!}{
\begin{tabular}{l|c}
 \textbf{Parameter} & \textbf{Value } \\ \hline
 Number of epochs & 3 \\
 Global batch size &  128*8 \\
 Optimizer &  Adam \\
 Maximum learning rate &  1e-4 \\
 Learning rate policy & Polynomial decay \\
 Warmup ratio &  0.1 \\
 Gradient norm clipping & 1.0 \\
 Max. input sequence length &  128 \\ 
 Dropout & 0.1\\
\end{tabular}}
\caption{\label{tab:hyper_parameters} Training hyper-parameters for BERT-base-cased and BERT-large-cased models}
\end{table}

\section{Dataset Statistics}\label{sec:data_stats}
\begin{table}[h]
\resizebox{\columnwidth}{!}{
\begin{tabular}{l|ccc}
 \textbf{Aspect}   & \textbf{Train}&\textbf{Dev}&\textbf{Test}  \\ \hline
    Overlapping services with training data & 1.0 & 0.43  & 0.22 \\
    Data amount of all tasks & 879k & 180k & 250k\\
    Data amount of each task (in \%)
    & 16:46:31:4:2 
    & 9:30:19:30:12 
    & 10:29:20:28:13 \\
    Negative sample ratio for each task (in \%)
    & 72:98:89:78:0
    & 65:98:89:98:0 
    & 67:98:88:97:0
\end{tabular}}
\caption{\label{tab:dataset} Dataset statistics on single-domain for the five tasks intent classification, slot request prediction, slot status prediction, value prediction and value extraction.}
\centering
\end{table}

\section{Test Results}\label{sec:test_results}
\begin{table}[h]
\centering
\resizebox{\columnwidth}{!}{
\begin{tabular}{l|cccc}
 \textbf{Model}   & \textbf{Intent Acc}&\textbf{Slot Req F1}&\textbf{Average GA} & \textbf{Joint GA} \\ \hline
 SGD baseline~\cite{rastogi2019towards} & 88.4(75.1/92.3) & 94.9(99.2/93.6) & 65.9(91.4/58.3) & 35.9(68.9/26.2) \\
 $\text{FastSGT}_{ST}$~\cite{noroozi2020fastsgt} & 88.3(73.9/92.6) & 94.6(99.2/93.2) & 70.4(95.3/62.9) & 39.3(81.5/27.0)  \\
 $\text{SGD-QA}_{ST}$ & \textbf{89.2(75.9/93.1)} & \textbf{98.8(99.6/98.6)} & \textbf{90.4(97.6/88.2)} & \textbf{67.9(91.5/61.0)} \\  \hline

\end{tabular}}
\caption{\label{tab:single_domain} Accuracy comparisons between the SGD baseline, FastSGT, and SGD-QA on \textbf{single-domain} test set for all services (seen/unseen). We excluded the configurations that are not available. Average of three runs are reported.}
\end{table}

\end{document}